\documentclass[letterpaper, 10 pt, conference]{ieeeconf}  

\usepackage{amsmath,amssymb,breqn}
\usepackage[ruled,vlined]{algorithm2e}
\usepackage{lipsum}
\usepackage[normalem]{ulem}
\useunder{\uline}{\ul}{}
\usepackage{makecell}\usepackage{color}
\usepackage{subfigure}
\usepackage{graphicx}
\usepackage[    colorlinks=true,
    linkcolor=blue,
    filecolor=blue,      
    urlcolor=blue,
    citecolor=blue,]{hyperref}

\IEEEoverridecommandlockouts                              

\title{\LARGE \bf
Edge Coverage Path Planning for Robot Mowing}

\author{Zhaofeng Tian$^{1}$,  and Weisong Shi$^{1}$  
\thanks{}
\thanks{$^{1}$CAR lab, Department of Computer Science,
        Wayne State University, Detroit, USA
        {\tt\small weisong@wayne.edu}}%
}

\begin{document}

\maketitle
\thispagestyle{empty}
\pagestyle{empty}

\begin{abstract}

Thanks to the rapid evolvement of robotic technologies, robot mowing is emerging to liberate humans from the tedious and time-consuming landscape work. Traditionally, robot mowing is perceived as a ``Coverage Path Planning'' problem, with a simplification that converts non-convex obstacles into convex obstacles. Besides, the converted obstacles are commonly dilated by the robot's circumcircle for collision avoidance. However when applied to robot mowing, an obstacle in a lawn is usually non-convex, imagine a garden on the lawn, such that the mentioned obstacle processing methods would fill in some concave areas so that they are not accessible to the robot anymore and hence produce inescapable uncut areas along the lawn edge, which dulls the landscape's elegance and provokes rework. To shrink the uncut area around the lawn edge we hereby reframe the problem into a brand new problem, named the ``Edge Coverage Path Planning'' problem that is dedicated to path planning with the objective to cover the edge. Correspondingly, we propose two planning methods, the ``big and small disk'' and the ``sliding chopstick'' planning method to tackle the problem by leveraging image morphological processing and computational geometry skills. By validation, our proposed methods can outperform the traditional ``dilation-by-circumcircle'' method. \url{https://sites.google.com/view/cutedge}

\end{abstract}


\section{INTRODUCTION}
With the fast evolution of the robotic landscape, more people are expected to be liberated from the time-costing and repeating tasks like indoor delivery, floor cleaning, and lawn mowing, wherein lawn mowing is particularly exhausting for many families. Currently, several mowing robot products are on the market, giving customers a glimmer of hope to be liberated from such labor work. The requirement of the robot mowing is to cover the lawn as much as possible within a reasonable time and energy cost. However, the design of the current mowing robots is limited by cost assessment hence the robot relies on the preset boundary signal wires or signal bases and plans a random path with collision sensors to stop itself, which makes the robot hardly perceived as intelligent to fulfill the mentioned requirement. To make mowing robots distinguished from general household appliances and promote the popularization of robot mowing, more sensing and planning technologies are tended to be employed.

Commonly considered a branch field of mobile robots, research on robot mowing is scoped from several similar perspectives to mobile robots, like mapping, planning, and control, however, some problems are unique in robot mowing.  For instance, when mapping or constructing the environment of a lawn, the commonly used mapping sensor, Lidar, does not help a lot to recognize the lawn boundaries to define a work region. Instead, camera-based lawn edge detection methods are widely surveyed by studies~\cite{ed1,ed3}. Additionally, path planning for a robot lawn mower mainly focuses on how to cover the lawn instead of planning a feasible path from A to B in the traditional navigation problems, thus the path planning for the robot mowing is considered a Coverage Path Planning (CPP) problem in most previous studies ~\cite{zong,zong2}. 

Generally, a CPP problem cares about how a robot can move through all the waypoints in the target area with the objective to optimize path length and energy or time costs. More specifically, to solve a CPP problem, it is usually converted into an optimal path finding or traversal problem in a discrete map, wherein the obstacles are usually assumed as convex, e.g., a circle, a triangle, and a polygon to simplify the environment construction~\cite{zong}. Additionally, a non-convex obstacle will be converted into a convex obstacle, and the convex obstacles are usually dilated by the robot's circumcircle to make sure the robot would not collide with them.~\cite{li, convex}.  

The above obstacle processing methods work when the workspace is large, e.g., for an agricultural machine to cover a farm field~\cite{field}, where the work field is much larger than the obstacle area. However, when considering robot mowing within a relatively smaller workspace shown in fig.1, the situation is quite different. Hence we find two issues when applying the traditional obstacle processing of the CPP problem to the lawn mowing scenario. 

\begin{figure}[htbp]
\centering
\subfigure[Boundary 1]{
\begin{minipage}[t]{0.48\linewidth}
\centering
\includegraphics[width=1.63in]{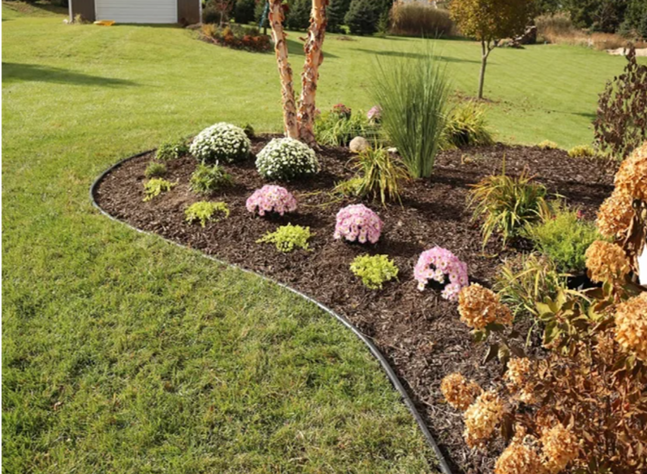}
\end{minipage}%
}%
\subfigure[Boundary 2]{
\begin{minipage}[t]{0.48\linewidth}
\centering
\includegraphics[width=1.6in]{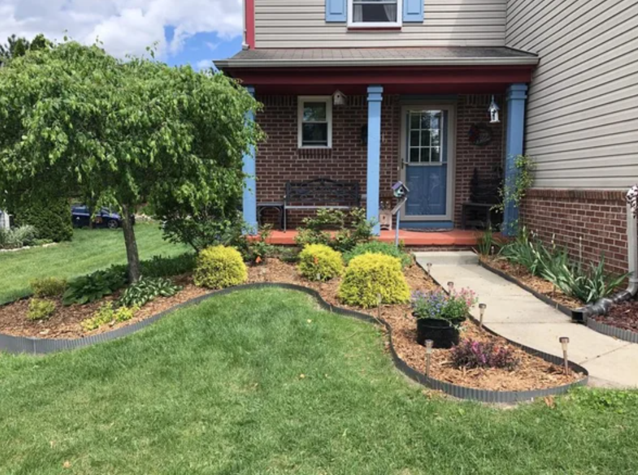}
\end{minipage}%
}%
\centering
\caption{Mowing boundaries, commonly seen in the garden of the single-family house, which consists of convex and concave sections.  }
\end{figure}

First, the obstacle boundary (also the lawn boundary, where the obstacle could be considered the garden area, tree area, or some areas else that are inaccessible to the robot) is curvy and comprises some concave ``valleys'', hence converting the non-convex obstacle into a convex polygon that could contain the obstacle, would inevitably fill in these ``valleys'' and make them inaccessible to the robot, which produces the uncut area. 

Second, the commonly used dilation method will increase the distance between the robot mowing deck and the lawn boundary, which would produce more uncut area along the boundary. In Fig.~2, an equivalent characterization of the dilation method is intuitively illustrated within a lawn mowing context.
 
 \begin{figure}[htbp]
\centering
\subfigure[Dilation-by-circumcircle]{
\begin{minipage}[t]{0.48\linewidth}
\centering
\includegraphics[width=1.62in]{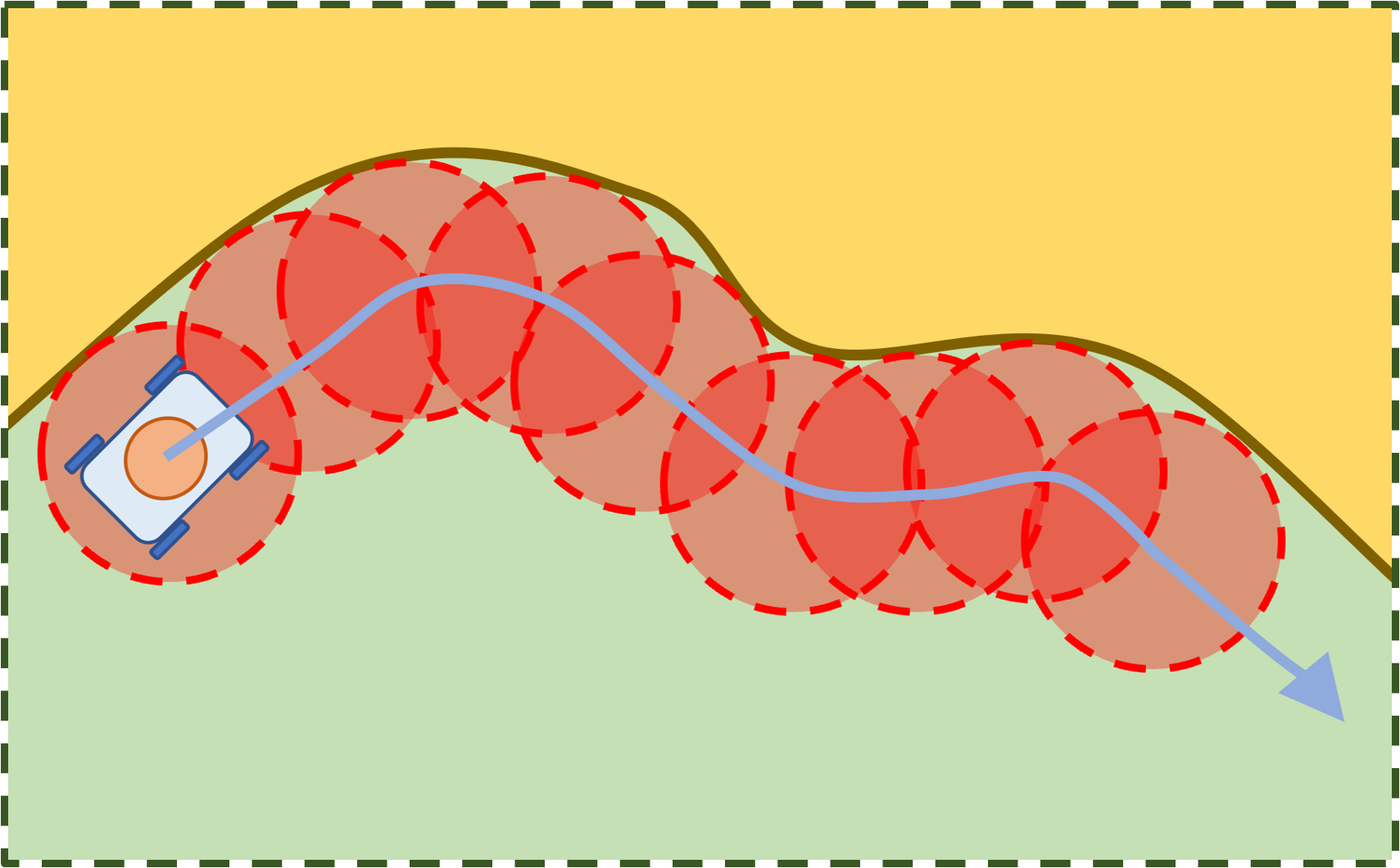}
\end{minipage}%
}%
\subfigure[Uncut area]{
\begin{minipage}[t]{0.48\linewidth}
\centering
\includegraphics[width=1.6in]{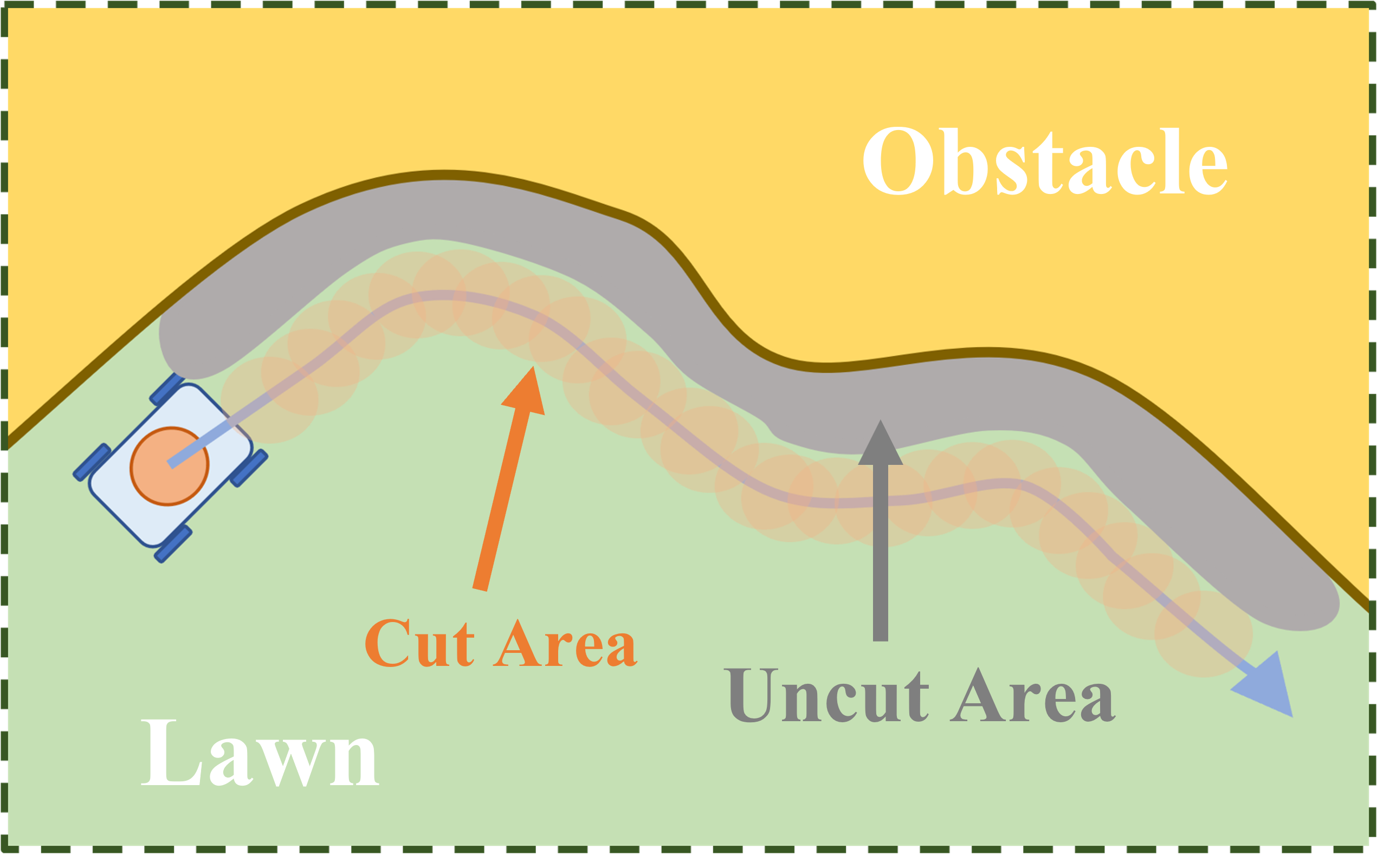}
\end{minipage}%
}%
\centering
\caption{Dilation and uncut area, by using the dilation method, the obstacle will be dilated by a disk that shares the same radius with the robot's circumcircle~(``dilation-by-circumcircle''). In (a), we depict the ``dilation-by-circumcircle'' method in an equivalent way, where the robot is contained by its circumcircle~(also the dilating disk shown as the red disk) and moving along the original obstacle boundary~(imagine that the circumcircle is rolling along the lawn boundary), hereby the trajectory of its center equivalently forms the boundary of the dilated obstacle~(blue line). As a result, it produces the uncut area shown as the grey area in (b) while the orange area is the area cut by the robot's mowing deck (shown as the orange disk inside the robot) because the dilation increases the distance between the robot's mowing deck to the original obstacle/lawn boundary.}
\end{figure}

\textbf{Our contributions.} \emph{(i) Edge~Coverage Path Planning problem.}~To tackle the under-coverage issue in the edge area which is overlooked within the CPP narrative environment, we hereby reframe the issue under a novel problem - the Edge Coverage Path Planning~(ECPP) problem. The proposed brand new ECPP problem is dedicated to the robot mowing scenarios and carefully defined by the essential parameters of both the robot and the landscape information, whose objective is to produce as smaller an uncut area as possible instead of traversal all waypoints in a CPP problem. 

\emph{(ii) Edge Coverage Path Planning methods.}~ Correspondingly we propose two ECPP methods to tackle the ECPP problem, wherein one is ``big and small disk'' planning, and the other is  ``sliding chopstick'' planning. The two methods leverage image morphological processing and computational geometry to plan a collision-free for differential-drive mowing robot so as to produce a smaller uncut area along the lawn edge area. By validation, the proposed methods can outperform the traditional ``dilate-by-circumcircle'' method.

\section{RELATED WORK}
\subsection{Coverage Path Planning for Robot Mowing}
With the development of autonomous system technologies, coverage operations are intended to be autonomous to save repeating and time-consuming labor work. Research~\cite{field} presents a full coverage path planning method for a car-like grazing robot to cover the whole ranch and investigates the effect of the front-wheel steering rate on the coverage path planning performance. ~\cite{07coverage} proposes an algorithm that takes the kinematic constraints of an Ackermann-steering into account to optimize the path curvatures as well as avoid outdoor obstacles. Research~\cite{mow1} designs an algorithm for a mowing robot, which  takes the field boundaries, polygon-based obstacle information, and kinematic constraints as input, and outputs a coverage path that is smoothed by Dubin's curve. While~\cite{mow2} employs double heuristic optimization algorithms to plan an optimal path for mowing robots, wherein the obstacles are denoted by disks, rectangles, and polygons. Besides,~\cite{mow3} proposes an ``p-d'' circuitous planning algorithm for a robot mower to cover a rectangular work region, wherein ``p-d'' two letters simulate the trajectory curves. Previous studies leverage the CPP method to solve the mowing problem in a macro scope, while they usually
consider the obstacles as circles and polygons, which are convex. But in real-world lawns, boundaries are not always convex. Also, due to the dilation operation, the CPP method could leave uncut areas along the boundaries.

\subsection{Lawn Edge Detection}
As some research indicates, setting up a border wire around the lawn to define the lawn boundaries is time-consuming and boring. Thus, recognition and detection of the lawn boundaries via computer vision technologies become an interesting topic to improve the efficiency of robot mowing.
Study~\cite{ed1} leverages the Gabor filters and Support Vector Machine (SVM) to recognize the lawn boundaries for mowing robots. ~\cite{ed2} designs a lawn boundary detection device using a line infrared light source and a camera with a bandpass infrared filter, which can work for various grass terrains. ~\cite{ed3} proposes a method to distinguish cut and uncut lawn areas by measuring grass height through a photo-interrupter sensor. ~\cite{ed4} proposes a Grey-Level Co-occurrence Matrix (GLCM) based method to differentiate cut and uncut grass and uses a Genetic Algorithm (GA) to optimize the parameters in the GLCM. These studies enable boundary extraction and cut-uncut area differentiation and could complement our proposed planning methods in terms of acquiring landscape information.


\section{EDGE COVERAGE PATH PLANNING PROBLEM}
In this section, we define the Edge Coverage Path Planning (ECPP) problem in detail. 
\subsection{Overall Formulation}
The workspace is defined as $\mathbb{O}\subset{\mathbb{R}^2}$, wherein the obstacle space, and the lawn space or called free space are defined as $\mathbb{O}_{obs}$, $\mathbb{O}_{lawn}\subset{\mathbb{O}}$. For the ECPP problem, the initial condition is defined by the boundary $\textbf{D} =\left\{ (x,y)  \mid (x, y) \in \{\mathbb{O}_{lawn}\cap{\mathbb{O}_{obs}} \}\right\} $, which separates the obstacle and the lawn area. The objective of the planning methods is to plan a path in the edge area $  \textbf{P} = \{(x, y) \mid (x, y) \in \mathbb{O}_{lawn} \}$, which could leave as the less uncut area within the edge area as possible, whereby the edge area or simply called edge $ \mathbb{O}_{edge}\subset{\mathbb{O}_{lawn}}$, can be defined as the area between the planned path $\textbf{P}$ and the boundary $\textbf{D}$. While the cut area $\mathbb{O}_{cut} \subset{\mathbb{O}_{edge}}$, is defined as the area between the planned path and the upper boundary of the area that is swept by the mowing robot, where the upper boundary denotes the boundary close to the lawn boundary. And the uncut area $\mathbb{O}_{uncut} = \mathbb{O}_{lawn} \ominus \mathbb{O}_{cut}$, can be defined as the unswept area left within the edge area. Thus, A Solution to an ECPP problem, or called an ECPP method, can be considered as aimed to plan a path $\textbf{P}$ to minimize the area value of the uncut area $\mathbb{O}_{uncut}$ under the given initial condition, i.e., the boundary $\textbf{D}$. Moreover, these definitions are explained in Fig. 3. as well.

\begin{figure}[htbp] 
\centering 
\includegraphics[width=0.6\columnwidth]{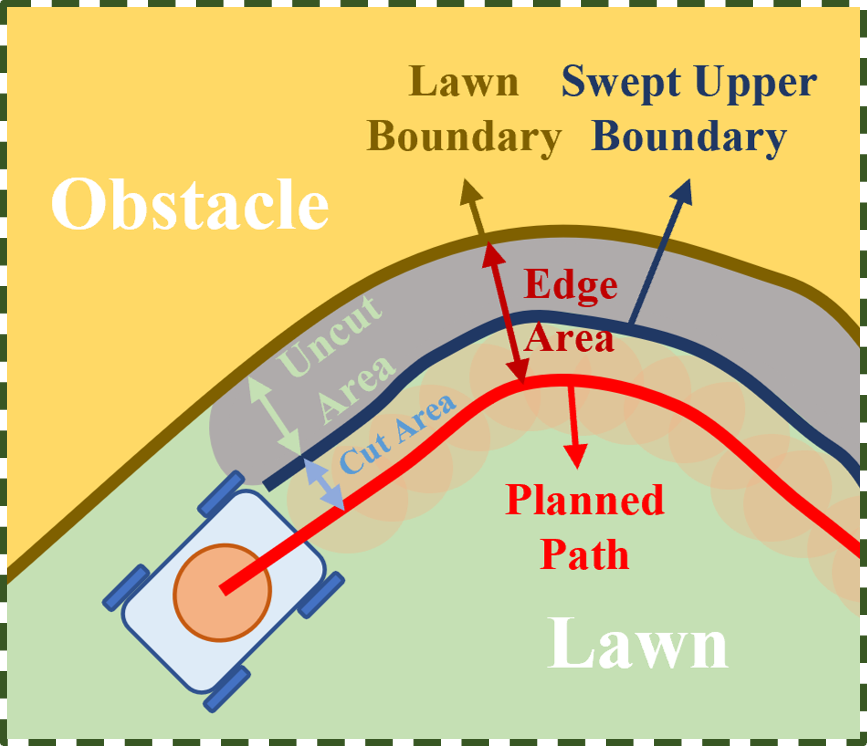} 
\caption{Different areas, this figure explains different areas and boundaries that are defined in the Edge Coverage Path Planning~(ECPP) problem.} 
\label{real} 
\end{figure}

\subsection{Problem Constraints}
To specify the constraints that are applied to the problem, the working condition of the mowing robot needs to be further explained. The robot is considered a differential-drive robot, the most common type seen among the products. And it is subject to the general non-holonomic kinematics.

Besides, the robot is considered bilateral and front-to-back symmetric, the center of the robot is also the center of the mowing desk. And the center of the robot is expected to move on the planned path with a corresponding heading angle at each moment. Additionally, the body of the robot should be out of the obstacle boundary.

\section{Methods}
To tackle the ECPP problem, we propose two image- morphology-based methods, the ``big and small disk'' planning, and the ``sliding chopstick'' planning.
\subsection{Morphological Image  Processing}
Some preliminary knowledge of our proposed method is illustrated here. Firstly, the leveraged techniques, dilation, and erosion in the image morphology are introduced \cite{gonza}.

\begin{equation}
    \hat{\textbf{B}} = \{ w \mid w= -b, b \in \textbf{B} \}
\end{equation}
\begin{equation}
    (\textbf{B})_z = \{ c \mid c= b+z,b\in\textbf{B} \}
\end{equation}
\begin{equation}
    \textbf{A} \oplus \textbf{B} = \{ z \mid \lbrack (\hat{\textbf{B}})_z \cap \textbf{A} \rbrack \subseteq \textbf{A} \}
\end{equation}
\begin{equation}
    \textbf{A} \ominus \textbf{B} = \{ z \mid (\textbf{B})_z \cap \textbf{A}^c = \emptyset \}
\end{equation}

 In equations (1) and (2), $\hat{\textbf{B}}$ is the set of the points in $\textbf{B}$ whose $(x,y)$ coordinates have been replaced by $(-x,-y)$, and $(\textbf{B})_z$ denotes the set after the translation of $\textbf{B}$ by point $z = (z_1, z_2)$. As equation (3) represents that the dilation of $\textbf{A}$ by a Structural Element (SE) $\textbf{B}$ is the set of all displacements, $z$, such that the foreground elements of $\hat{\textbf{B}}$ overlap at least one element of $\textbf{A}$. Erosion of $\textbf{A}$ by a SE $\textbf{B}$ is expressed in equation (4), which denotes the set of all points $z$ such that $B$, translated by $z$, is contained in A, to another word, does not share any common elements with the background, i.e., the complement set of $\textbf{A}$, $\textbf{A}^c$.
 \begin{equation}
     \textbf{A} \circ \textbf{B} = (\textbf{A} \ominus \textbf{B}) \oplus \textbf{B}
 \end{equation}
\begin{equation}
     \textbf{A} \bullet \textbf{B} = (\textbf{A} \oplus  \textbf{B}) \ominus \textbf{B}
 \end{equation}
Besides, opening and closing techniques are expressed as equations (5) and (6). The opening $\textbf{A}$ by a SE $\textbf{B}$ is the erosion of $\textbf{A}$ by $\textbf{B}$, and then dilated by $\textbf{B}$. While the closing $\textbf{A} $ by a SE $\textbf{B}$ is the dilation of $\textbf{A}$ by $\textbf{B}$, followed by erosion of the result by $\textbf{B}$.

\subsection{Robot and Boundary Parameters}
To simulate the real lawn boundary, we assume the boundary $\textbf{D}= \{(x,y) \mid y= \textit{f}(x)\}$ is a smooth continuous curve, where each point $(x,y)$ in the curve is subject to a bijection mapping relation $y=f(x)$. Moreover, as shown in Fig.~4, the length $l$ and the width $w$ are the length and width of the robot body. And $R_1 = (l/2)^2 + (w/2)^2$ is the radius of the ``big disk'', which is the circumcircle of the rectangular-shaped robot and is considered the safety region of the robot, where the differential robot can rotate in situ without interfering with outside objects. In addition, the ``big disk'' is also the disk that the obstacle is dilated by in the traditional ``dilation-by-circumcircle'' method. $R_2 = w/2$ is the radius of the ``small disk'', which is the inscribed circle of the robot. While $R_3$ is the radius of the ``mowing disk'', which is the mowing deck and the working area of a mowing robot. These three disks are centered on the center of the mowing robot. 
\subsection{Boundary Preprocessing}
To avoid the robot being stuck in some narrow non-convex section of the boundary curve, the raw boundary needs to be preprocessed. By using the ``big disk'' as an SE to run the closing process on the boundary, a new boundary after the process can be obtained. The new boundary will fit the ``big disk'' in the new non-convex section and make sure the robot would not be stuck inside the ``narrow valley'' since now the robot can rotate inside the `` big disk '' within the new boundary to adjust the direction instead of being stuck and then backing up. 

\subsection{``Big and Small Disk'' Planning}
The reason for the name is that the ``big disk'' with $R_1$ and the ``small disk'' with $R_2$ introduced above are used to do the morphological image processing. Simply speaking, the main idea of this proposed method is to dilate the preprocessed boundary by the ``big disk'' in the non-convex sections and by the ``small disk'' in the convex sections. Intuitively, suppose we want to minimize the uncut area in the ECPP problem. In that case, the mowing disk is expected to be as close to the boundary as possible where the boundary indicates the preprocessed boundary. Ideally, if we use a ``small disk'' to dilate the boundary, equivalently the robot's body side can tangent to the curve while moving, which makes the distance between the ``mowing disk'' closest to the boundary at the distance of $R_2-R_3$. However, this works only in the convex sections, in the non-convex sections, the part of the robot body would violate the boundary. Hence this ``small disk'' planning is perceived as imaginary. To compensate for the violation issue, in the non-convex sections, the safety disk, i.e. the ``big disk'' will be used to dilate the preprocessed boundary so that driving long the dilated boundary will not violate the preprocessed boundary.

\begin{figure}[htbp]
\centering
\subfigure[Parameters]{
\begin{minipage}[t]{0.47\linewidth}
\centering
\includegraphics[width=1.45in]{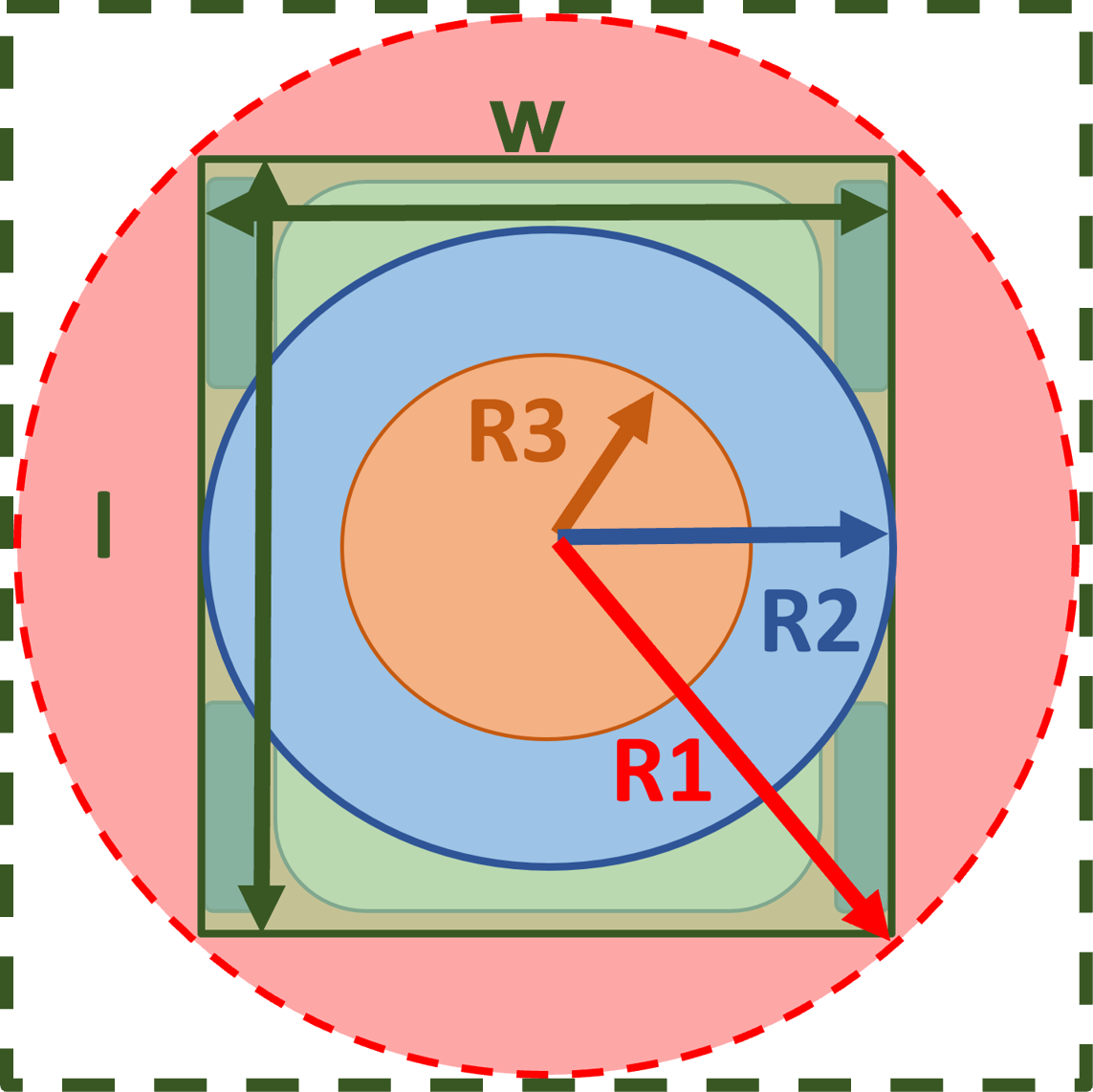}
\end{minipage}%
}%
\subfigure[Boundary preprocessing]{
\begin{minipage}[t]{0.47\linewidth}
\centering
\includegraphics[width=1.6in]{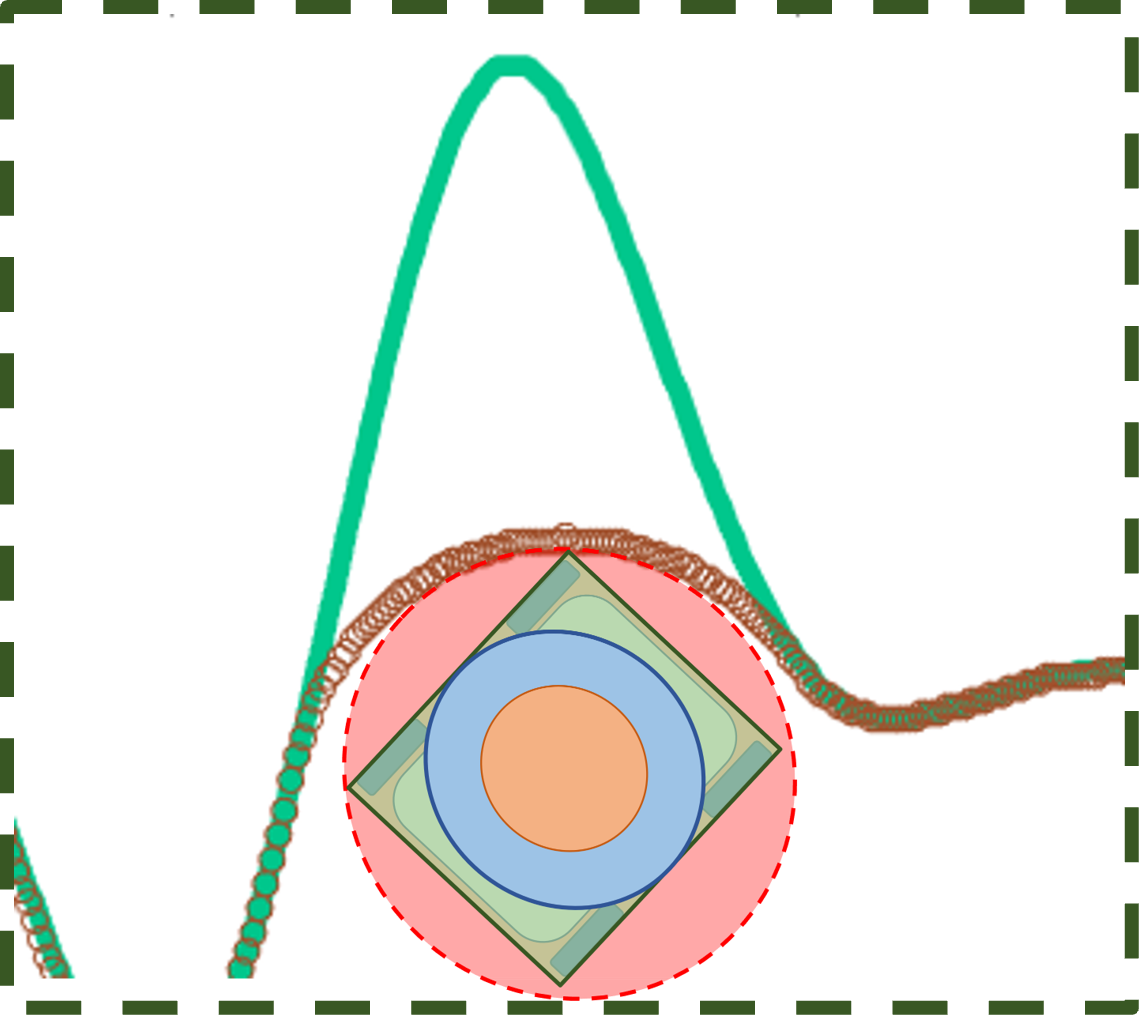}
\end{minipage}%
}%
\centering
\caption{ Parameters and boundary preprocessing, wherein the ``big disk'' with radius $R_1$, is used to operate the morphological closing process on the boundary. ``Small disk'' with radius $R_2$ and ``mowing disk'' with radius $R_3$ respectively denote the inscribed circle and the mowing deck of the robot.}
\end{figure}

\begin{algorithm}[!ht]
\DontPrintSemicolon
\SetKwInput{KwInput}{Input}                
\SetKwInput{KwOutput}{Output}              
\SetKwFunction{FBSDP}{BSDP}
\SetKwFunction{FMain}{Main}
\SetKwFunction{FCollision}{Collision}
  \KwInput{raw boundary $\textbf{D}$, big disk $\textbf{E}_{b}$, small disk $\textbf{E}_{s}$}
  \KwOutput{planned path $\textbf{P}$}
  \SetKwProg{Fn}{Def}{:}{}

\Fn{\FBSDP{\textbf{D}, $\textbf{E}_{b}$, $\textbf{E}_{s}$ }}
{
    $\textbf{P}, \textbf{Y}_{path} \leftarrow $ [ ] \;
    $\textbf{D}^*(\textbf{X}, \textbf{Y})\leftarrow$ $preprocess( \textbf{D}) $ \;
    $\Ddot{\textbf{Y}} \leftarrow$ second-order derivative of $\textbf{Y}$\;
    $\textbf{Y}_{big} \leftarrow$ lower bound of $\textbf{D}^* \oplus \textbf{E}_b$\;
    $\textbf{Y}_{small} \leftarrow$ lower bound of $\textbf{D}^* \oplus \textbf{E}_s$\;

    \For{$i \leftarrow 1$ \KwTo $len( \textbf{X})$}
    {   
        \If{$\Ddot{\textbf{Y}}[i] > 0$}
        {
           $ \textbf{Y}_{path}[i] \leftarrow \textbf{Y}_{small}[i]$
        }
        \Else 
        {
             $ \textbf{Y}_{path}[i] \leftarrow \textbf{Y}_{big}[i]$
        }
        
    }
    $\textbf{P} \leftarrow (\textbf{X},\textbf{Y}_{path}$) \;
    \KwRet{$\textbf{P}$}

}

\caption{Big and Small Disk Planning}
\end{algorithm}
In Algorithm 1, the preprocessed boundary is obtained as $\textbf{D}^*$ which consists of two coordinate vectors $\textbf{X}$ and $\textbf{Y}$. Then the second-order derivative of $\textbf{Y}$, i.e., $\Ddot{\textbf{Y}}$ indicates whether the curve is convex or concave at the point $(x_i,y_i), x_i \in \textbf{X}, y_i \in \textbf{Y}$. Following two vectors $\textbf{Y}_{big} $ and $\textbf{Y}_{small}$ are respectively obtained from dilating $\textbf{D}^*$ by the ``big disk'' $\textbf{E}_b$ and the ``small disk'' $\textbf{E}_s$, and extracting the lower bounds of both dilated area. Next, iterate $\Ddot{\textbf{Y}}$, wherein $\Ddot{\textbf{Y}}[i] > 0$ indicates the point is convex and choose $\textbf{Y}_{small}[i]$  to be inserted in $\textbf{Y}_{path}[i]$. Finally, $(\textbf{X},\textbf{Y}_{path}) $ is assigned to the planned path $\textbf{P}$. 

\subsection{Sliding Chopstick Planning}

A chopstick sliding in a bowl resembles a robot moving along the concave boundary to reach more coverage, hereby we call this proposed method ``sliding chopstick'' planning shown in Fig.5. As we mentioned in the ``big and small disk'' planning, in the convex section, the robot's ``small disk'' can be moving tangent to the boundary so that reaches more coverage in the concave section, while the ``big disk'' can move along the boundary when it is concave. However, using the big disk to move along the boundary will definitely keep the robot safe but compromise the uncut area. Consequently, we propose the ``sliding chopstick'' planning method to tackle this problem that let the robot's one side slide in the concave section like a chopstick sliding in a bowl, so that leaves as less uncut area as possible without violating the boundary. The details can be expressed as Algorithm 2.

\begin{figure}[htbp] 
\centering 
\includegraphics[width=0.87\columnwidth]{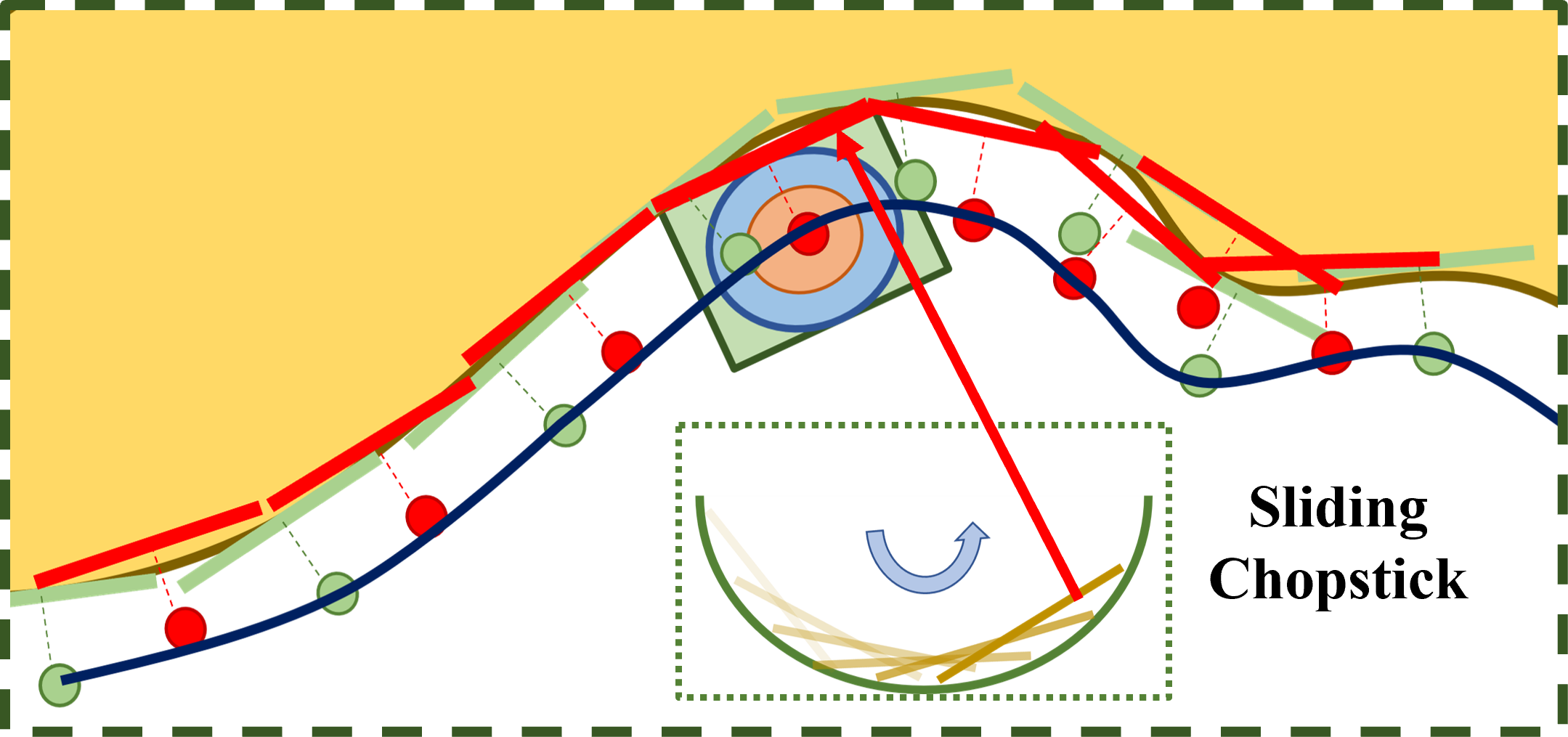} 
\caption{Sliding Chopstick Planning, the red line, and red point denote the robot side and robot center in the ``sliding chopstick'' planning, while the green line and point denote the ``small disk'' planning. In the convex section, the robot centers of ``small disk planning'' are lower than the ``sliding chopstick planning'', and in the concave section, conversely higher, where the robot's side simulates a chopstick sliding in a bowl. And the lower centers from both methods will construct the final path~(blue line).} 

\label{real} 
\end{figure}

\begin{algorithm}[!ht]
\DontPrintSemicolon
\SetKwInput{KwInput}{Input}                
\SetKwInput{KwOutput}{Output}              
\SetKwFunction{FSCP}{SCP}

  \KwInput{raw boundary $\textbf{D}$, robot length $l$, width $w$, small disk $\textbf{E}_{s}$}
  \KwOutput{planned path $\textbf{P}$}
  \SetKwProg{Fn}{Def}{:}{}

\Fn{\FSCP{\textbf{D}, $\textbf{E}_{s}$, $l$, $w$ }}
{
    $\textbf{P}, \textbf{Y}_{path},\textbf{Y}_{slide} \leftarrow $ [ ] \;
    $\textbf{D}^*(\textbf{X}, \textbf{Y})\leftarrow$ $preprocess( \textbf{D}) $ \;
    $\Ddot{\textbf{Y}} \leftarrow$ second-order derivative of $\textbf{Y}$\;
    $\textbf{Y}_{small} \leftarrow$ lower bound of $\textbf{D}^* \oplus \textbf{E}_s$\;
    
    \For{$i \leftarrow 1$ \KwTo $len( \textbf{X})$}
    {
        \For{$j \leftarrow i$ \KwTo $len( \textbf{X})$}
        {
            \If{$dist((x_i,y_i), (x_j, y_j )) = l $}
            {
                $line \leftarrow ((x_i,y_i),(x_j, y_j))$\;
                find the point $(a,b)$ that: \;
                $dist((a,b), line)= \frac{w}{2} $ \;
            }
            
        }
        $\textbf{Y}_{slide}[i] \leftarrow b$
    }

    \For{$i \leftarrow 1$ \KwTo $len( \textbf{X})$}
    {
        \If {$\textbf{Y}_{slide}[i] < \textbf{Y}_{small}[i]$}
        {
            $\textbf{Y}_{path}[i] \leftarrow \textbf{Y}_{slide}[i]$
        }
        \Else
        {
            $\textbf{Y}_{path}[i] \leftarrow \textbf{Y}_{small}[i]$
        }
    }

    $\textbf{P} \leftarrow (\textbf{X},\textbf{Y}_{path}$) \;
    \KwRet{$\textbf{P}$}
}
\caption{Sliding Chopstick Planning}
\end{algorithm}

\begin{figure*}[!htb] 
\centering 
\includegraphics[width=0.87\textwidth]{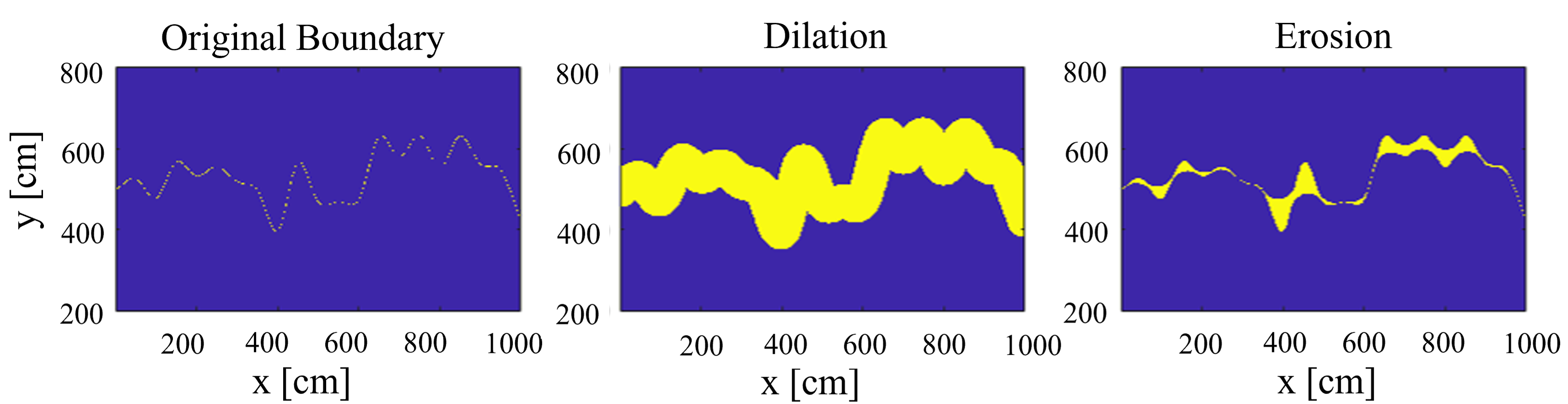} 
\caption{Boundary Preprocessing, from left to right, the original boundary is first dilated and then eroded by the ``big disk'', which constitutes the morphological closing process.} 
\label{real} 
\end{figure*}

In Algorithm 2, two vectors of $y$ coordinate are created, wherein $\textbf{Y}_{small}$ is the lower bound of the dilated boundary by the ``small disk'', and $\textbf{Y}_{slide}$ is obtained from the ``sliding chopstick'' method. In the ``sliding chopstick'' method, given a $(x_i,y_i)$ point in the preprocessed boundary, it finds another $(x_j,y_j)$ point in the boundary that makes the distance between two points equal to the length of the side of the robot body, based on which the coordinates of the robot center can be found and recorded into $\textbf{Y}_{slide}$. And since the most drastic bumping changes in the original boundary are removed by the boundary preprocess, we can consider within the robot length, the boundary would not vary its convexity more than once. Hereby, in the convex section, the ``sliding chopstick'' will be inside of the obstacle, and conversely in the concave section, will be out of the obstacle. And for ``small disk'' planning, the robot will be outside of the lawn boundary when it is convex and will violate the boundary when it is concave. So combing the ``sliding chopstick'' method and ``small disk'' planning, at a given $x_i$, we choose the one $y_i$ either from $\textbf{Y}_{slide}$ or $\textbf{Y}_{small}$ so that whether the boundary is convex or concave we always choose a lower point from the two methods to avoid violation.

\section{Experiments and Results}
Further, we validate our planning methods in the simulation, then the path obtained from the ``sliding chopstick'' planning is tracked in the real-world test.  
\subsection{Planning Results}
In the simulation, the original boundary is simulated by the triangle function, which is shown in Fig.~6. From the left to right, firstly the yellow line represents the original boundary,  the part upper than the line represents the obstacle and the lower part represents the lawn. Then, to preprocess the boundary, the closing process is operated on the original boundary, which consists of two steps, first dilation and then erosion. In the second subpicture, the yellow area denotes the boundary dilation by the ``big disk'', which is the circumcircle of the robot. Finally, in the last subpicture, the dilated area is eroded by the ``big disk'' again, leaving the ``valley'' areas filled up. Hereby, the preprocessed boundary can be extracted from the lower bound after the closing process.
\begin{figure}[!htb] 
\centering 
\includegraphics[width=\columnwidth]{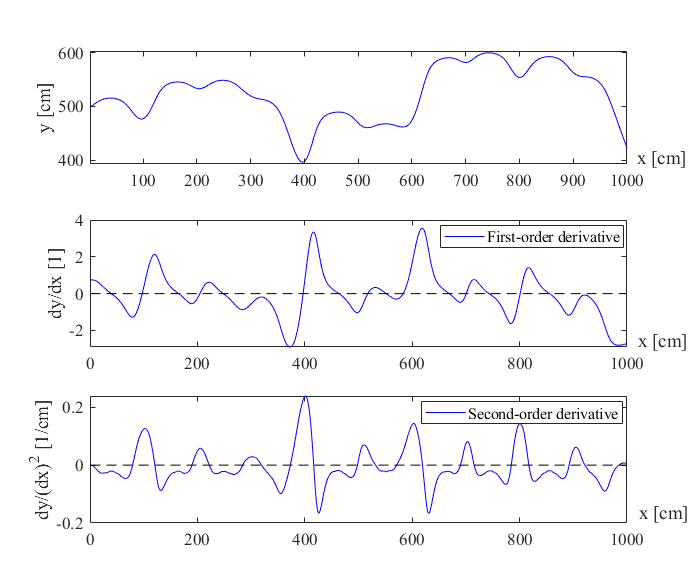} 
\caption{Convexity calculation, from the top to bottle, the first subpicture shows the boundary after preprocessing, followed by the first and second-order derivative calculations, wherein the second-order derivative represents the convexity, which is positive when convex and negative when concave.   } 
\label{real} 
\end{figure}

\begin{figure}[!htb] 
\centering 
\includegraphics[width=1\columnwidth]{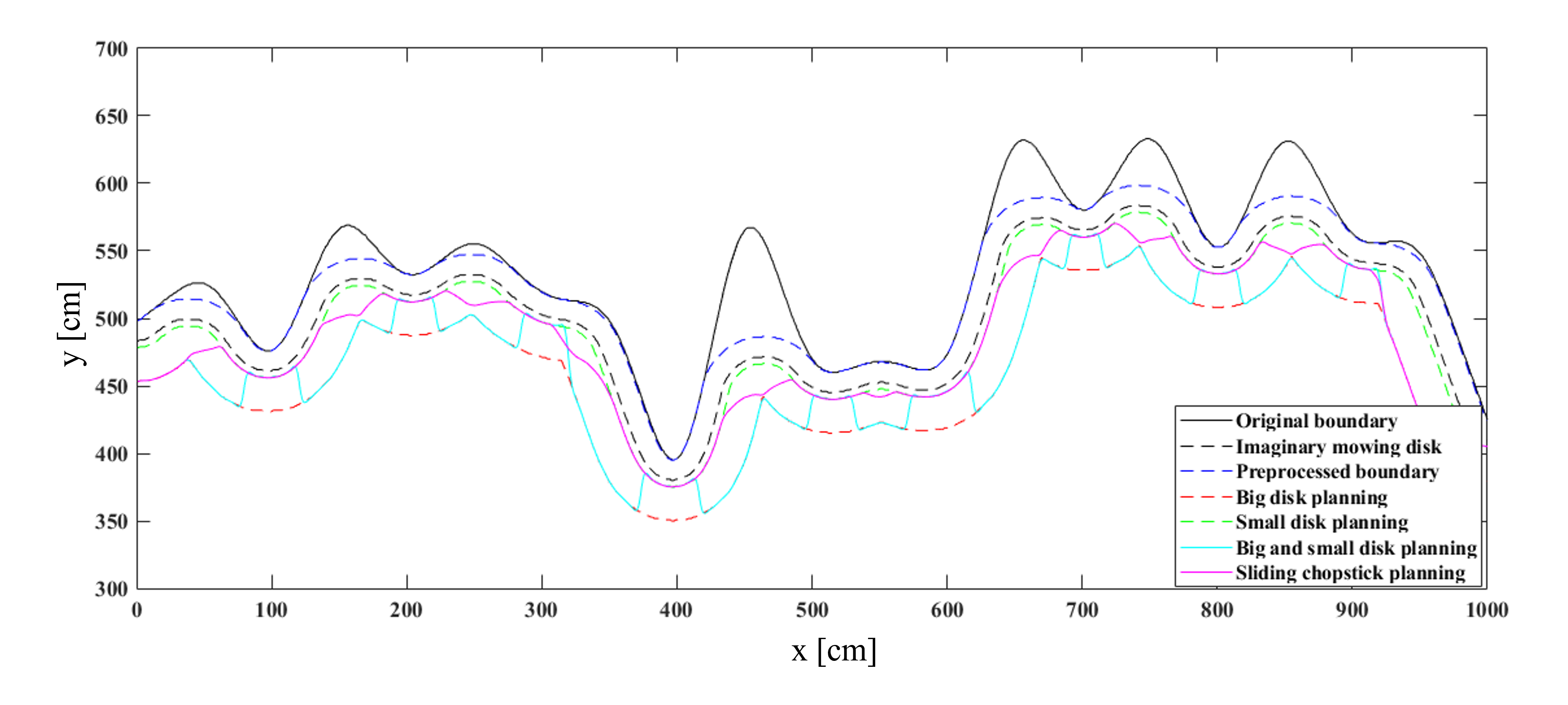} 
\caption{Planning results comparison, the proposed ``big and small disk'' planning and ``sliding chopstick'' planning can reach more coverage than the ''big disk'' planning~(``dilation-by-circumcircle''). } 
\label{real} 
\end{figure}

Then the first-order and second-order derivatives of the preprocessed are calculated in Fig.~7, wherein the second-order derivative denotes the convexity of the preprocessed boundary. Subsequently, the planned paths from imaginary ``mowing disk'', ``small disk'', ``big disk'', ``big and small disk'', and ``sliding chopstick'' planning methods are plotted in Fig.~8. Notice that the imaginary ``mowing disk'' and ``small disk'' serve only as references to the proposed methods and actually could cause violations in the real world. Additionally, the paths of ``big and small disk'' and ``sliding chopstick'' planning are smoothed by splines to compensate for the robot's kinematic stability.
\begin{figure}[h] 
\centering 
\includegraphics[width=0.95\columnwidth]{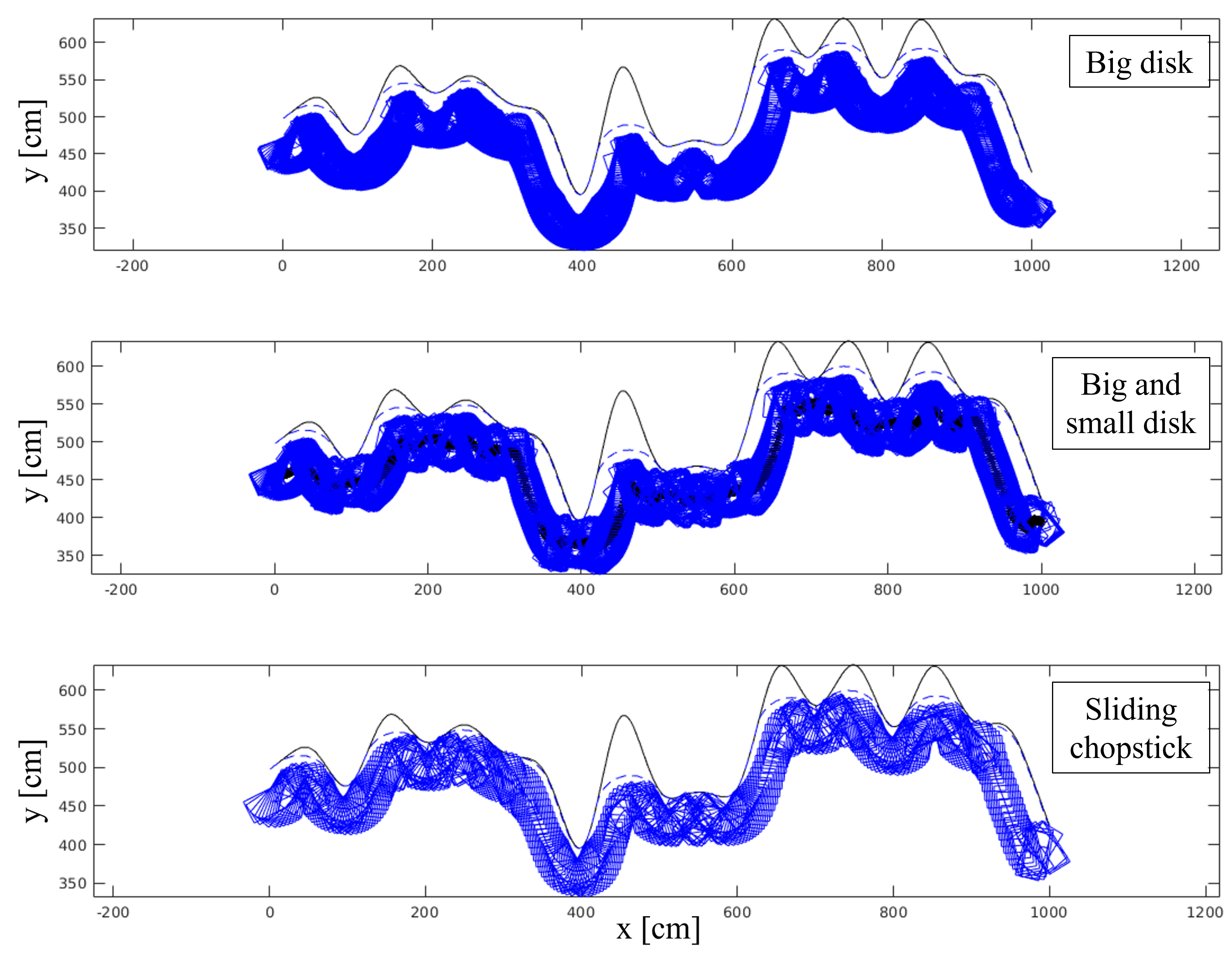} 
\caption{Robot trajectories, the trajectories of the robot obtained from ``big disk'', ``big and small disk'', and ``sliding chopstick'' planning are plotted in order, which demonstrates the planned path is collision-free. } 
\label{real} 
\end{figure}

As shown in Fig.~9, the trajectories of the ``big disk'', ``big and small disk'', and ``sliding chopstick'' planning are respectively plotted from top to bottom. Where the parameters of the robot are selected as $l=0.8$~$m$, and $w=0.4$~$m$, and the resolution of the discrete map is 0.01 meter. The robot body trajectories are subject to boundary limits within an error range of 0 to 0.02~$m$ caused by resolution settings. 

\begin{figure}[htbp] 
\centering 
\includegraphics[width=0.76\columnwidth]{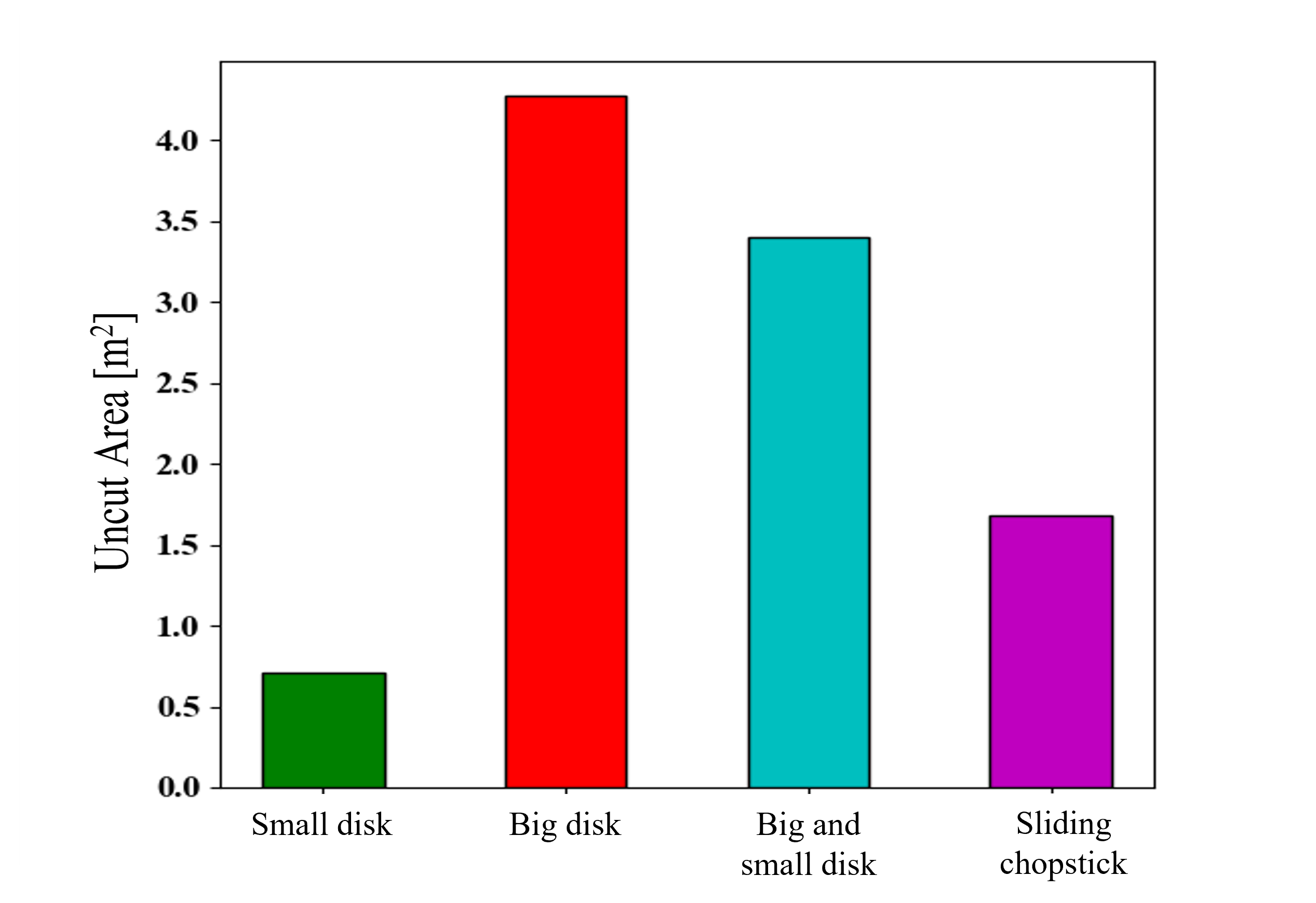} 
\caption{Uncut area comparison, where the result of the ``small disk'' planning is imaginary and serves as a reference.} 
\label{real} 
\end{figure}

\begin{figure}[htbp] 
\centering 
\includegraphics[width=0.5\columnwidth]{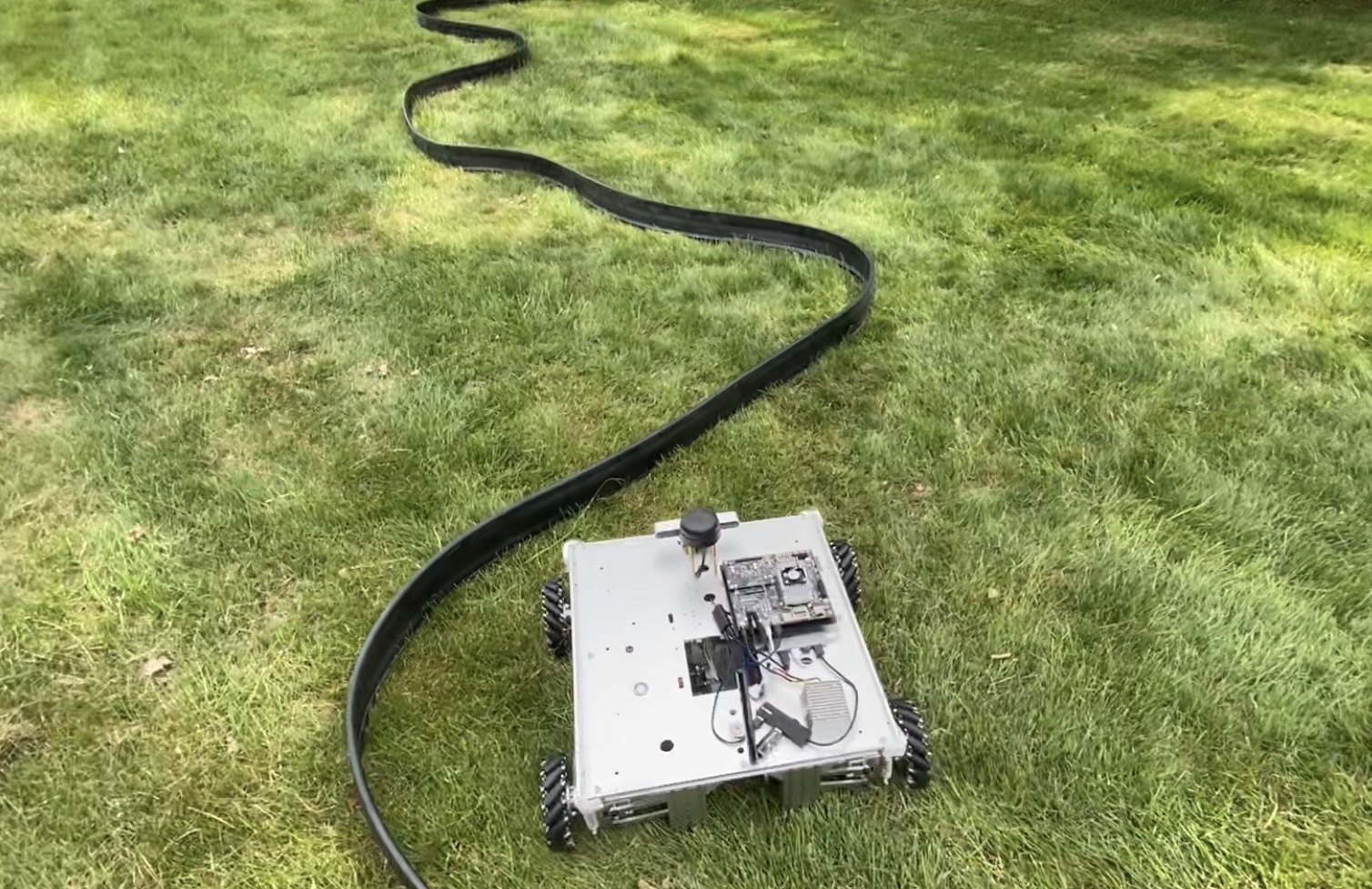} 
\caption{Real-world path tracking test, by using simple proportional controllers we test the path tracking with a differential robot.} 
\label{real} 
\end{figure}


To numerically analyze the performance of the proposed methods in the context of the defined ECPP problem, the value of the uncut areas are compared in Fig.~10, given the ``mowing disk'' with a radius of $R_3 =0.15$~$m$. As we can see, the imaginary ``small disk'' planning produces the least uncut area of $0.718$~$m^2$, while the ``big disk'' planning produces the most uncut area of $4.2734$~$m^2$. Our proposed ``big and small disk'' and ``sliding chopstick'' planning produce uncut areas of $3.4039$~ $m^2$ and $1.676$~$m^2$, which are respectively $20.35\%$ and $ 60.78\% $ smaller than the ``big disk'' planning, wherein the ``sliding chopstick'' planning can produce a satisfying result that approaches the idealized ``small disk'' planning.
\subsection{Real-world Tracking Test} 
To investigate the feasibility of the proposed methods, the path obtained from the ``sliding chopstick'' planning is tracked by a differential Mecanum-wheel robot shown in Fig.~11 using simple proportional controllers. Wherein, the black edging roughly simulates the lawn boundary and the robot localization is  based solely on the odometry. Hereby, the final result is plotted in Fig.~12, where the real path is denoted by the blue line while the reference path is denoted by the green line. The existing errors are considered reasonable given that the lawn is slippery and uneven, which detriments the precision of odometry, and further the performance of localization. Note that the contribution of this study is concentrated on the planning part while the path tracking only forms a sketchy demonstration.

\begin{figure}[htbp] 
\centering 
\includegraphics[width=0.9\columnwidth]{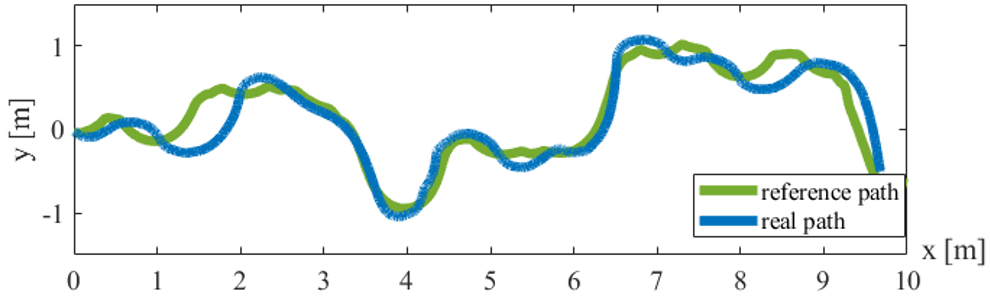} 
\caption{Path tracking result, reference path obtained from the ``sliding chopstick'' planning is shown as the green line, while the real tracking path is shown as the blue line. } 
\label{real} 
\end{figure}

\section{Conclusions and Discussions}
By simulation, our proposed ``big and small disk'' and ``sliding chopstick'' planning methods can output an appropriate collision-free path for the mowing robot to move along the lawn edge, meanwhile, fulfill a better performance than the ``big disk'' planning (traditional ``dilation-by-circumcircle'' method) by producing smaller uncut areas in our proposed Edge Coverage Path Planning~(ECPP) problem context. Furthermore, we find, slippery grass, and controller design would generate uncertainties in the practice. By the limits of this study, the proposed solutions are merely hoped to be a cornerstone for more in-depth studies on the ECPP problem from other researchers in the near future.

\bibliographystyle{IEEEtran}
\bibliography{rm}

\end{document}